\documentclass[letterpaper]{article} 
\usepackage{aaai23}  
\usepackage{times}  
\usepackage{helvet}  
\usepackage{courier}  
\usepackage[hyphens]{url}  
\usepackage{graphicx} 
\usepackage{natbib}  
\usepackage{caption} 
\usepackage{algorithm}
\usepackage{algorithmic}

\usepackage{newfloat}
\usepackage{listings}
\DeclareCaptionStyle{ruled}{labelfont=normalfont,labelsep=colon,strut=off} 
\floatstyle{ruled}
\newfloat{listing}{tb}{lst}{}
\floatname{listing}{Listing}
\usepackage{microtype}
\usepackage{graphicx}

\usepackage{xcolor}
\usepackage{amsmath,amssymb}
\usepackage{amsmath}
\usepackage{lipsum}
\usepackage{subcaption}
\usepackage{comment}
\usepackage{url}
\usepackage{multirow}
\usepackage{adjustbox}                          
\usepackage{booktabs}
\DeclareUnicodeCharacter{2212}{-}

\nocopyright

\title{Weakly Supervised Learning for Analyzing Political Campaigns on Facebook}
\author {
    Tunazzina Islam\footnote{Corresponding author. Email: islam32@purdue.edu}, 
    Shamik Roy,
    Dan Goldwasser  \\
}
\affiliations {
    Department of Computer Science, Purdue University\\
    West Lafayette, Indiana 47907 \\
   \{islam32, roy98, dgoldwas\}@purdue.edu
}

\begin{document}

\maketitle

\begin{abstract}


Social media platforms are currently the main channel for political messaging, allowing politicians to target specific demographics and adapt based on their reactions. However, making this communication transparent is challenging, as the messaging is tightly coupled with its intended audience and often echoed by multiple stakeholders interested in advancing specific policies. Our goal in this paper is to take a first step towards understanding these highly decentralized settings. We propose a weakly supervised approach to identify the stance and issue of political ads on Facebook and analyze how political campaigns use some kind of demographic targeting by location, gender, or age. Furthermore, we analyze the temporal dynamics of the political ads on election polls.
\end{abstract}

\section{Introduction}
 
Over the last decade, social media has impacted public discourse and communication, particularly in the political context \cite{kushin2010did,wattal2010web,ratkiewicz2011detecting,stieglitz2013social,jensen2017social,marozzo2018analyzing,8508646,ferrara2020characterizing,sharma2021characterizing}.
Social media has a transformative effect on how political candidates interact with potential voters by adapting their messaging to different demographic groups' specific concerns and interests. This process, known as microtargeting~\cite{hersh2015}, relies on data-driven campaigning techniques that exploit the rich information collected by social networks about their users. By measuring the users’ engagement with political content, candidates can identify the issues, and even the specific phrases and slogans, that resonate with each demographic group. Furthermore, political campaigns on social media are highly distributed, with multiple stakeholders and interest groups using the platforms to advance their interests and show support for different candidates by specifically focusing on agenda items relevant for their interests (e.g., the National Rifle Association (NRA) might emphasize the track-record of each candidate on protecting gun rights).  

Our goal in this paper is to take a first step towards analyzing and monitoring the landscape of political advertising on social media. We focus our experiments on the U.S. 2020 presidential elections, analyzing content supporting either the Biden-Harris or the Trump-Pence campaigns. Our goal is twofold: first to characterize the different stakeholders and analyze their content, and second to build on this characterization to analyze political messaging across different demographics. 
We deal with the decentralized nature of political advertising on social media. We analyze over $5K$ advertisers (referred to as \textit{funding entities}) that funded over $800K$ political ads on Facebook\footnote[2]{\url{https://www.facebook.com/ads/library/api}}, associating advertisers with a binary label (Pro-Biden or Pro-Trump) and ads with four categories capturing positive or negative messaging and its target. We also identify the specific policy issue discussed in the ad, a 13-class classification problem. To clarify, consider the following two ads: 
\newline
\textbf{Ad1:} \textit{\small From COVID-19 to the environment to racial justice, Donald Trump has failed. Joe Biden and Kamala Harris can set us on a new course. The stakes for Pennsylvanians could not be higher.}
\newline
\newline
\textbf{Ad2:} \textit{\small President Trump PROTECTED Social Security and Medicare. Joe Biden tried to cut them MULTIPLE times. President Trump LOWERED drug costs, and Medicare Advantage Premiums fell 34\%. Under Biden, drug prices SKYROCKETED. Joe Biden and Kamala Harris's FAR left plan threatens private insurance and limits choices.}
\newline
Ad1 has Anti-Trump and Pro-Biden stances focusing on the multiple issues \textit{covid, climate, racial justice}. On the other hand, Ad2 has Pro-Trump and Anti-Biden stances focusing on the \textit{healthcare} issue.
\begin{figure}[htbp]
  \centering  
  \includegraphics[width= .47\textwidth]{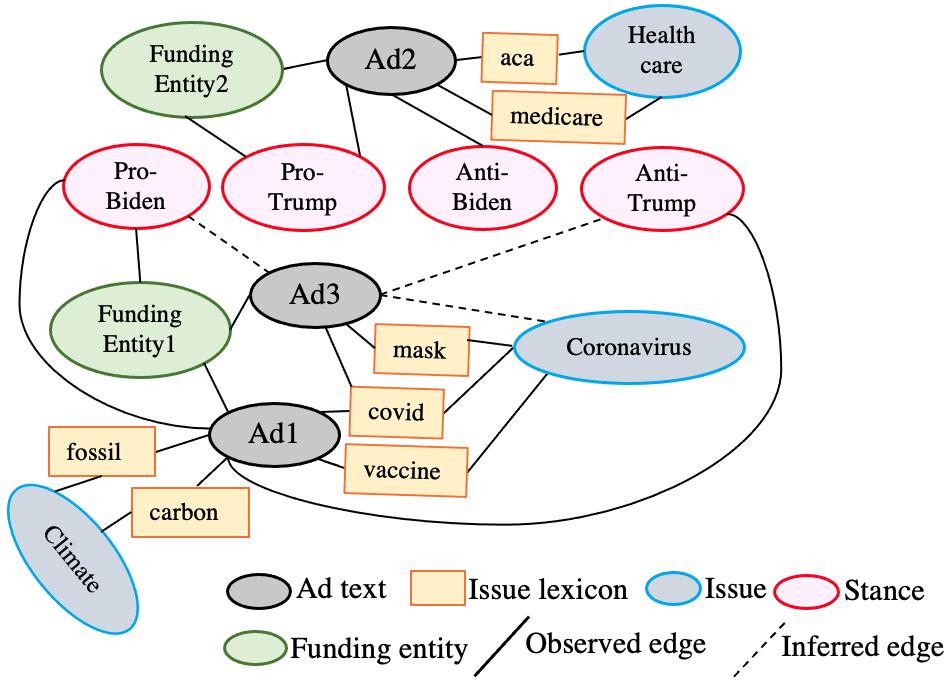}
    \caption{ Political advertising graph capturing relations among ads, funding entities, stances, issues, issue lexicons. A funding entity connects to Ad1, which has both pro-biden' and `anti-trump' stances, on issues related to `climate' and `coronavirus'.}
    \label{fig:graph}
\end{figure}

In this paper, we suggest a weakly supervised graph-embedding based framework in which ads and advertisers are learned jointly. While some cases, the name of the advertisers capture their bias, e.g.,
\textit{{\small `BIDEN FOR PRESIDENT', `TRUMP MAGA COMMITTEE'}} and we mention them as \textit{explicit} advertisers. Some advertisers \textbf{do not} explicitly mention any candidate name/party affiliation in their names, e.g.,
\textit{`Union 2020', `Plains PAC'} and we call them \textit{implicit} advertisers. We leverage weak supervision from explicit advertisers for the stance prediction. Our embedding objective is derived from dedicated lexicons developed for identifying policy issues, and by identifying the political position of a small number of advertisers (i.e., the position of the Trump-Pence campaign is known). During learning, our model learns the associations between advertisers and their positions and the content they publish, as well as the issues they are mostly concerned with. 

Fig. \ref{fig:graph} represents the embedding graph of our framework containing nodes i.e., \textbf{ads, funding entities, stances, issues, issue lexicons} and edges representing their relationships, i.e. as Ad1 has Pro-Biden stance, there is an edge between node \textbf{Ad1} and \textbf{Pro-Biden}. We learn a graph embedding to maximize the similarity between neighboring nodes \cite{perozzi2014deepwalk,tang2015line,grover2016node2vec}. Fig. \ref{fig:graph} shows that \textbf{Ad1} and \textbf{Ad2} have weak labels for stances and issues obtaining from their funding entities and issue lexicons respectively. Initially, \textbf{Ad3} doesn't have a label. It has edge with \textbf{Funding Entity1} and issue lexicons named `covid' and `mask'. 
We exploit these resources to train
our embedding model, which captures the context and, as a result, can generalize and grasp stances and issues in new ads (i.e., \textbf{Ad3} in Fig. \ref{fig:graph}). 

The learned model allows us to analyze how political candidates and stakeholders micro-target specific demographics. In this work, we examine a novel dataset of advertisements posted to Facebook during the 2020 U.S presidential election and make the full dataset available to the research community. Using the information provided by Facebook Political Ads API, we analyze the issues used for supporting (and attacking) each candidate on ads targeting different geographical regions, age, and gender groups. We evaluate the quality of the learned model by applying it to detect the stance of implicit funding entities and compare it with their views (ground truth). Further, we discuss how election polls affect political campaigns using Granger causality \cite{granger1988some}. 
We focus on the following research questions (RQ) to analyze political campaigns:
\begin{itemize}
    \item RQ1. Can we analyze political campaigns without direct supervision? (Section \textbf{Results and Analysis})
    
    
    \item RQ2. Are messages distinctive in ads? (Subsection \textbf{Descriptive Insights})
    \item RQ3. Which demographics are reached by advertisers? (Subsection \textbf{Audience Demographics}) 
    
     \item RQ4. How specific region is reached by advertisers and their messages? (Subsection \textbf{State-wise Issue and
Demographics})

    
     \item RQ5. Are election polls represented in ad campaigns? (Subsection \textbf{Granger Causality with Polls})

     

    
\end{itemize}

Our contributions are summarized as follows:
\begin{enumerate}
    \item We formulate a novel problem of exploiting weak supervision to analyze the landscape of political advertising on social media.
    \item We propose a weakly supervised graph embedding based framework to identify political stance of advertisers as well as the published content and issues of the content. We show that our model outperforms the baselines. 
    \item We conduct quantitative and qualitative analysis on real-world dataset to demonstrate the effectiveness of our proposed model.
\end{enumerate}
Our code and data are publicly available {here}\footnote[3]{\url{https://github.com/tunazislam/weaklysup-FB-ad-political}}.

\section{Related Work}
During elections, political candidates use social media for their campaigns. Recent works show monitoring and analysis of targeted advertising on social media \cite{andreou2019measuring,silva2020facebook,serrano2020political}. \citet{islam2022covidfbAd} analyzed Covid-19 vaccine campaign on Facebook. \citet{ribeiro2019microtargeting} analyzed political ads on Facebook that
are linked to a Russian propaganda group: Internet Research
Agency (IRA). \citet{silva2020facebook_ufl} showed Facebook ad (created by IRA) engagement targeting the 2016 U.S. general election. \citet{silva2020facebook} designed a system to monitor political ads on Facebook in Brazil and deployed during the Brazilian 2018 elections. \citet{capozzi2020facebook,capozzi2021clandestino} examined advertising concerning the issue of immigration in Italy. 
Our paper detects the stance and issue of political ads on Facebook. It analyzes political campaigns for both candidates based on the target audience's demographic and geographic information as well as presents temporal analysis for the 2020 U.S presidential election.

Recent works frame the issue of perspective detection as a text categorization problem~\cite{greene2009more, klebanov2010vocabulary, recasens2013linguistic, iyyer2014political, johnson2016all}. It is typically studied as a supervised learning task~\cite{lin2006side,durant2006predicting,greene2009more}. 
In contrast, our approach relies on weak supervision and lexicon based approaches~\cite{roy2020weakly, field2018framing}. Weakly supervised methods reduce dependence on labeled texts. Graph based semi-supervised algorithms achieved considerable attention over the years \cite{Zhu02learningfrom,belkin2006manifold,subramanya2008soft,talukdar2008weakly,sindhwani2008document,yang2016revisiting,hisano2018semi}. \citet{tang2015pte,zhang2020minimally} used graph-based methods to build text networks with words, documents and labels and propagate labeling information along the graph via embedding learning. \citet{han2016partially} encoded weakly supervised information in positive unlabelled learning tasks into pairwise constraints between training instances imposing on graph embedding.
Recently, \citet{islam2021twitter} proposed weakly supervised graph embedding based EM-style framework to characterize user types on social media. 
Our embedding model is similar to contrastive learning-based embedding \cite{wu2020clear,giorgi2020declutr}. However, contrastive learning is self-supervised, where labels are generated from the data without any manual or weak label sources. In our case, we generate the label using weak supervision.
Our work is also closely related to the entity-targeted sentiment analysis~\cite{mohammad2016semeval, field2019entity, mitchell2013open, meng2012entity}. In our work, we use weak supervision to identify stance and issue of political ads and analyze political campaigns. To the best of our knowledge, this is the first work to utilize a weakly supervised graph embedding based framework to analyze political campaigns on social media.

\section{Data}
We collect around $0.8$ million political ads from January-October 2020 using Facebook Ad Library API with the search term `biden', `harris', `trump', `pence'. All advertisements are written in English.
For each ad, the API provides
the ad ID, title, ad body, and URL, ad creation time and the time span of the campaign, the Facebook page authoring the ad, funding entity, the cost of the
ad (given as a range). The API also provides information on the users who have seen the ad (called `impressions'): the total number of impressions (given as a range and we take the average of the end points of the range), distribution over impressions broken down by gender (male, female, unknown), age ($7$ groups), and location down to states in the USA. 
We have duplicate content among those collected ads because the same ad has been targeted to different regions and demographics with unique ad id. We have $35327$ 
ads with different contents, $5431$ unique funding entities, among them $537$ explicitly mention candidate names and/or party affiliations, e.g., \textit{\small {BIDEN FOR PRESIDENT, DONALD J. TRUMP FOR PRESIDENT, INC}}. 

\subsection{Holdout Data}
For validation purpose, we manually annotate $667$ ads for stances and issues. We consider $4$ stances `pro-biden', `pro-trump', `anti-biden', `anti-trump' and $13$ issues\footnote[4]{\url{https://ballotpedia.org/}} called `abortion', `covid',  `climate', `criminal justice reform, race, law \& order', `economy and taxes', `education', `foreign policy', `guns', `healthcare', `immigration', `supreme court', `terrorism', `lgbtq'. We also mark `non-stance', `non-issue' ads. Two annotators from the Computer Science department manually annotate a subset of ads to calculate inter-annotator agreement using Cohen’s Kappa coefficient \cite{cohen1960coefficient}. This subset has inter-annotator agreements of $77.50\%$ for stance and $69.60\%$ for the issue, which are substantial agreements. In case of a disagreement, we resolve it by discussion. The rest of the data was annotated by one graduate student from the Computer Science department.

\section{Methodology}
We represent political advertising activity on social media as a graph, connecting funding entities to their ads. We represent the outcome of our analysis, stance and issue predictions, as separate label nodes in the graph connected via edges to ads and funding entities. Each issue label-node is associated with an $n$-gram lexicon, a set of nodes representing lexical indicators for the issue. Based on  known associations between funding entities and stances, we associate $10\%$ of the funding entities and their ads with stance labels. The lexicon and observed stance relations act as a weak form of supervision for  graph embedding. Our model learns to generalize the stance predictor to new ads, and by contextualizing the lexicon $n$-grams based on their occurrence in ads, we learn to associate other ads with the relevant issue even when the lexicon items are not present.
These settings are described in Fig. \ref{fig:graph}. Note that each ad can be associated with multiple issues and stances (e.g., pro-biden and anti-trump). 
   
    
    
    
    
    
    
    


   

\begin{table}[t]
\small
    \centering
    \resizebox{\columnwidth}{!}{%
    \begin{tabular}{c|c}
    \toprule
    \textbf{\textsc{Issue(uni,bi,tri)}} & \textbf{\textsc{Issue(uni,bi,tri)}} \\
    \midrule
    Abortion (56, 20, 1) &  Foreign policy (95, 31, 6) \\
    Covid (52, 23, 5) & Guns (92, 20, 6)\\
    Climate (66, 22, 3) & Healthcare (62, 21, 4)\\
    Criminal justice reform, race & Immigration (78, 25, 3) \\ law \& order (93, 26, 5) & Supreme court (80, 25, 4) \\
    Economy \& taxes (41, 16, 2) & Terrorism (73, 19, 3) \\
    Education (62, 22, 2) & LGBTQ (55, 12, 1)\\
    \bottomrule
    \end{tabular}}
    \caption{Number of unigram, bigram, trigram in each issue.}
    \label{tab:issue_lex}
\end{table}

\subsection {Issue Lexicon}
To create the issue lexicon, we collect $30$ news articles covering each issue from left leaning, right leaning, and neutral news media. We know the news source bias
from \url{https://mediabiasfactcheck.com/}. We calculate the Pointwise Mutual Information (PMI) \cite{church1990word} to identify issue-specific lexicons. We calculate the PMI for an $n$-gram, $w$ with issue, $i$ as $PMI(w,i) = log \frac {P(w|i)} {P(w)}$.
To compute $P(w|i)$, we take all news articles related to an issue $i$ and compute $\frac {count(w)} {count(all\_ngrams)}$. We have $30$ news articles per issue. $P(w)$ is computed by counting $n$-gram, $w$ over the whole corpus ($390$ news articles). 
We assign each $n$-gram to the issue with the highest PMI and build an $n$-gram lexicon for each issue. 
Table \ref{tab:issue_lex} shows the number of unigrams, bigrams, and trigrams with PMI $\geq 0.5$ per issue. 
In this paper, we use only unigrams, resulting in $905$ issue-indicating words. 

\subsection {Model}
To identify stances and issues, we do the followings:
\newline
\noindent\textbf{Inferring Stance Labels Using Knowledge.} 
In some cases, the names of funding entities capture their bias. For example: 
\textit{`Biden Victory Fund', `Keep Trump in office'} clearly state their position.
We extract all funding entities mentioning the candidates or their party. In addition, if the funding entity name also includes the words \textit{\{`dump', `lie', `out', `fail', `against'\}}, we assume the funding agency has stance against that candidate. In this manner, we annotate $537$ funding entities. We annotate the ads generated by those funding entities. If a `pro-trump' funding entity addresses `biden', the ad is `anti-biden' (and the vice-versa). We use this approach to provide labels for $5343$ ads, and use them as weak supervision for our model.
\begin{table}
\centering
\begin{tabular}{lcc}
    \toprule
    \textbf{Model}
     & \textbf{Accuracy}  & \textbf{Macro-avg F1}    \\
     \midrule
    BiLSTM\_Glove  &  0.54  & 0.42  \\
    Fine-tuned BERT & 0.56  & 0.41  \\
    Rule-based  &  0.38  &  0.33 \\
    \textbf{Our Model} &  \textbf{0.73} & \textbf{0.63}  \\
    \bottomrule
    \end{tabular}%
\caption{Model performance for the stance prediction task.} 
\label{tab:performance}
\end{table}
\newline
\noindent\textbf{Prediction of Stances and Issues of Ads and Funding Entities}. We embed the following instances in a common embedding space - (a) Ads (b) Funding Entities; (c) Issue Lexicon; (d) Issue Labels; (e) Stance Labels. We maximize the similarity between two instances in the embedding space if --
(1) A funding entity has a stance.~
(2) An ad has a stance.~
(3) An ad has a word from the issue lexicon.~
(4) An issue lexicon word has an annotated issue.
We follow a negative sampling approach to learn the embeddings. Given an instance $o$, a positive example $m^{p}$ and a negative example $m^{n}$, where $o$ is placed closer to $m^{p}$ and far from $m^{n}$ in the embedding space, 
the embedding loss designed to place $o$ closer to $m^{p}$ than $m^{n}$ is $ E_{r}(o,m^{p},m^{n})=l(\operatorname{sim}(o, m^{p}), \operatorname{sim}(o, m^{n}))$
Here, $E_{r}$ defines the embedding loss for objective type $r$ (for example, ad to stance). Our goal is to maximize the similarity of a node embedding with a positive example and minimize the similarity with a negative example. We call a stance a positive example for an ad, if the ad has the stance, otherwise, the stance is called a negative example. We randomly sample $2$ negative examples for each instance (ad) for the objectives numbered (1) and (2) and $5$ negative examples for (3) and (4). The number of negative examples provided differs by type of objectives because there are multiple stances and issues for the same instance. For example, let's assume an ad has stances both `pro-biden' and `anti-trump' which are positive examples for the ad, and negative examples for that would be `pro-trump' and `anti-biden' stances -- this is the reason for choosing $2$ negative examples randomly for objectives numbered (1) and (2). If the ad has multiple words from the issue lexicon, that means the ad has multiple issues that are considered positive examples. The rest would be the negative examples for the ad. Therefore, we randomly pick $5$ negative examples for the objectives numbered (3) and (4) for our data. $sim()$ is the dot product and $l()$ is the cross-entropy loss, $ l(\operatorname{sim}(o, m^{p}), \operatorname{sim}(o, m^{n})) =-\operatorname{log}(\frac{e^{\operatorname{sim}(o, m^{p})}}{e^{\operatorname{sim}(o,m^{p})}+e^{\operatorname{sim}(o,m^{n})}})$
For all kind of objectives, we minimize the summed loss $\sum_{r \in R} \lambda_{r} E_{r}$, where $R$ is the set of all objective functions and $\lambda_{r}$ is the weight for objective function of type $r$. We initialize $\lambda_{r}=1$, for all.
\section{Experimental Setup}
In this section, we present the baselines to evaluate the effectiveness of our model and hyperparameter tuning.
\subsection{Baselines}
For baseline comparison, we use rule-based ad stance prediction, based on $4$ simple paraphrases \textit{``We support Donald Trump", ``We support Joe Biden", ``We do not support Donald Trump", ``We do not support Joe Biden"}. We predict the stance of all ads based on the similarity with the paraphrases. 
\newline
For the supervised baseline, we train a model using the weak labels assigned by the knowledge from explicit funding entities and predict the stance of the ads having no weak label (test data).
From the weakly labeled training data, we randomly choose $20\%$ data as the validation set. Our first supervised baseline is referred to as BiLSTM\_Glove model. We use $300d$ Glove word embeddings to obtain the ad embeddings and pass them to bidirectional LSTM \cite{hochreiter1997long}. For the second supervised baseline, we fine-tune the pre-trained BERT-base-uncased \cite{devlin2019bert} model. For the BERT's input, we tokenize the text using BERT's wordpiece tokenizer. For both supervised baselines, we use cross-entropy loss, $5$-fold cross-validation and report the average test result (accuracy and macro-avg F1 score) considering only the manually annotated ads as ground truth.
\begin{figure}
\begin{subfigure}{.5\columnwidth}
  \centering
  \includegraphics[width=\textwidth]{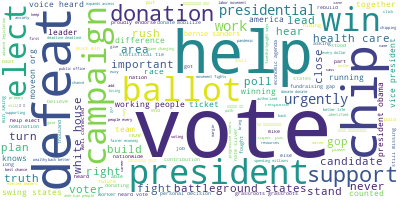}
  \caption{pro-biden}
  \label{fig:pro_biden}
\end{subfigure}%
\begin{subfigure}{.5\columnwidth}
  \centering
  \includegraphics[width=\textwidth]{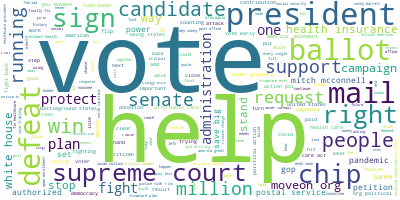}
  \caption{anti-trump}
  \label{fig:anti_trump}
\end{subfigure}
\begin{subfigure}{.5\columnwidth}
  \centering
  \includegraphics[width=\textwidth]{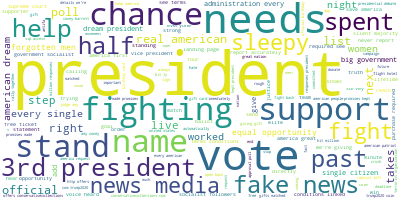}
  \caption{pro-trump}
  \label{fig:pro_trump}
\end{subfigure}%
\begin{subfigure}{.5\columnwidth}
  \centering
  \includegraphics[width=\textwidth]{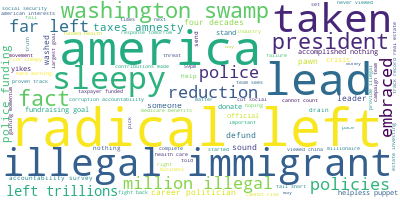}
  \caption{anti-biden}
  \label{fig:anti_biden}
\end{subfigure}
\caption{Wordcloud for stances. Importance of each word is shown with font size and color.
Pro-biden and anti-trump stances are talking about \textit{vote, ballot, chip}. Pro-trump has words like \textit{president, vote}. Anti-biden has noticeable words such as \textit{radical left, sleepy, illegal immigrant}.}
\label{fig:stance_wc}
\end{figure}
\begin{table*}
\begin{adjustbox}{width=\textwidth,center}
\begin{tabular}{lccc|lccc}
    \toprule
    \textbf{FE} & \textbf{GT}  & \textbf{Pred\_M}  & \textbf{Acc\_M (\%)} & \textbf{FE} & \textbf{GT}  & \textbf{Pred\_M}  & \textbf{Acc\_M (\%)}   \\
    \midrule
    Planned Parenthood Votes  & L  & L & 100.0 & Women Speak Out PAC  & C & C & 95.5 \\
    Union 2020  & L & L & 82.8 & America First Action  &  C & C &  98.3 \\
    United We Dream Action & L & L & 100.0  & PRESERVE AMERICA PAC  &  C & C & 100.0 \\
    Independence USA PAC  & L & L  & 99.9  & PROTECT FREEDOM PAC & C & C & 68.5  \\
    PACRONYM & L & L & 92.6 & American Potential Fund & C & C & 76.9 \\
    Alliance for a Better Minnesota & L & L & 98.4 & Plains PAC & C & C & 100.0 \\
    Black PAC & L & L & 100.0 & Citizens for Free Enterprise & C & C & 100.0\\
    The Lincoln Project & L & L & 100.0  & AG TOGETHER PAC & C & C & 100.0 \\
    DEFEAT DISINFO PAC & L & L & 93.9  & FREEDOM THROUGH TRUTH & C & C & 63.2 \\
    UNITED FOR PROGRESS PAC & L & L & 92.1 &  Family Policy Alliance  & C & C & 75.0 \\
    Veterans For Responsible Leadership & L & L & 100.0  &  RESTORATION PAC & C & C & 75.0 \\
    Dream Defenders Fight PAC  & L & L & 100.0 & \textbf{AMERICANS FOR PROSPERITY} &  \textbf{C} & \textbf{L} & \textbf{22.4} \\
    Working America & L & L & 100.0 & \textbf{Championing America at Her Best} & \textbf{C} & \textbf{L} & \textbf{-} \\
    UNITE THE COUNTRY & L & L & 100.0 & \textbf{C3 PAC} &  \textbf{C} & \textbf{L} & \textbf{46.6} \\
    FOR OUR FUTURE & L & L & 100.0 & Honoring American Law Enforcement PAC & C & C & 100.0\\
    BEST DAYS AHEAD & L & L & 100.0 & CONGRESSIONAL LEADERSHIP FUND & C & C & 72.5 \\
    REALLY AMERICAN PAC & L & L & 84.3 &  WE ARE GREAT AGAIN PAC & C & C & 72.9 \\
    Win Justice & L & L & 84.8 & KEEP AMERICA GREAT COMMITTEE & C & C & 73.2 \\
    VOTE VALUES & L & L & 82.0 & GREAT AMERICA PAC & C & C & 100.0 \\
    NARAL Freedom Fund & L & L & 90.9 & Keep Kentucky Great & C & C & 100.0 \\
    COMMON DEFENSE ACTION FUND & L & L & 100.0 & STOP SOCIALISM NOW PAC & C & C & 100.0 \\
    DEFEAT BY TWEET & L & L & 100.0 & Wisconsin Right to Life & C & C & 75.0 \\
    NJ7 CITIZENS FOR CHANGE & L & L & 100.0 & Texas Right to Life & C & C & 100.0 \\
    NEW POWER PAC & L & L & 100.0 & FLORIDA 8TH PAC & C & C & 100.0 \\
    QUESTION PAC & L & L & 100.0 & CatholicVote & C & C & 100.0\\
   \bottomrule
    \end{tabular} 
\end{adjustbox}
\caption{ Model performance on implicit funding entities for predicting stance. Bold funding entities have prediction error (see details in the Qualitative Analysis section). L= Liberal; C= Conservative; FE= Funding entity; GT= Ground truth stance of funding entity; Pred\_M= Predicted stance by our model; Acc\_M= Predicted accuracy by our model. }
\label{tab:result}
\end{table*}
\subsection{Hyperparameter Details}
Ad embeddings are obtained by running a Bi-LSTM \citep{schuster1997bidirectional} over the Glove \citep{pennington2014glove} word embeddings of the words of the ad. We concatenate the hidden states of the two opposite directional LSTMs to get representation over one time-stamp and average the representations of all time-stamps to get a single representation of the ad. All the embeddings are initialized in a $300d$ space. We use single layer Bi-LSTM which takes $300d$ Glove word embeddings as inputs and maps to a $150d$ hidden layer. We train this Bi-LSTM jointly with the embedding learning. Adam \cite{kingma2014adam} optimizer with learning rate $0.001$ is used. We initialize the embeddings of all of the other instances randomly. we use validation loss as a stopping criteria. We run the embedding learning at most $100$ epochs or stop the learning if the loss does not decrease for $10$ consecutive epochs.
For BiLSTM\_Glove baseline, the single layer Bi-LSTM takes $300d$ Glove word embeddings as inputs and maps to a $150d$ hidden layer with optimizer= $Adam$, learning rate= $0.01$, batch size = $32$, epochs = $20$. For BERT fine-tuning, we use maximum sentence length = $200$, batch size = $32$, learning rate = $2e-5$, optimizer= $AdamW$ \cite{loshchilov2018decoupled}, epochs = $3$, epsilon parameter = $1e-8$. We chose the lowest validation loss as a stopping criterion for the two supervised baselines.
\section{Results and Analysis}
\label{Results}
After learning the embeddings from our weakly supervised graph embedding approach, we can infer the stances and issues of unlabeled ads and funding entities by looking at the embedding similarity of the ads and funding entities having stance and issue labels. 
The stance or issue label having the maximum similarity is inferred as the predicted issues and stances. We discard `non-stance' and `non-issue' cases for the stance and issue prediction evaluation, resulting in $544$ ads and  $174$ ads, respectively. We use accuracy and macro-average F1 score as the evaluation metrics.
We obtain accuracy of $73.71\%$ and macro-average F1 score of $62.81\%$ for stance prediction of the ads. For the issue prediction tasks, we achieve $68.39\%$ accuracy and $54.76\%$ macro-average F1 score. 
For rule-based baseline of stance prediction we get $38.42\%$ accuracy and $33.17\%$ macro-average F1 score (Table. \ref{tab:performance}).  We compare our model with two supervised baselines. BiLSTM\_Glove obtains $53.55\%$ accuracy and $41.62\%$ macro-average F1 score for stance prediction (Table. \ref{tab:performance}). Second supervised baseline (Fine-tuned BERT) gets $55.91\%$ accuracy and $40.70\%$ macro-average F1 score (Table. \ref{tab:performance}). Table \ref{tab:performance} shows that weakly supervised graph embedding model achieves the highest performance in stance prediction compare to the rule-based as well as supervised baselines which answers the \textbf{RQ1}.


\begin{table*}
\begin{adjustbox}{width=\textwidth,center}
\begin{tabular}{llllllll}
    \toprule
      \multicolumn{2}{c}{\textbf{Pro-biden}} & \multicolumn{2}{c}{\textbf{Anti-trump}} & \multicolumn{2}{c}{\textbf{Pro-trump}} & \multicolumn{2}{c}{\textbf{Anti-biden}} \\
      \cmidrule(r){1-2}\cmidrule(l){3-4}\cmidrule(r){5-6}\cmidrule(l){7-8}
     \textbf{Trigrams}  & \textbf{P\_Cor}  & \textbf{Trigrams}  & \textbf{P\_Cor} & \textbf{Trigrams}  & \textbf{P\_Cor}  & \textbf{Trigrams}  & \textbf{P\_Cor} \\
    \midrule
    vote joe biden  &  0.763 & defeat donald trump  & 0.594 & president trump need & 0.667 & joe biden democratic & 0.713 \\
    presidential election held & 0.470  & request ballot today & 0.526 & trump need vote & 0.729 & dont let america & 0.357\\
    joe kamala democrat  &  0.579  &  time running out  & 0.238 & vote november $3^{rd}$ & 0.539 & taken joe biden & 0.600 \\
    today vote democrat  &  0.583  & not authorized candidate  & 0.353 & $3^{rd}$ president trump & 0.510 & radical left taken & 0.431\\
    vote democrat joe & 0.683  & affordable care act  & 0.382 & fake news medium & 0.303 & democratic party dont &  0.524\\
    defeat donald trump  &  0.535  & new trumpcare plan  & 0.572 & president trump spent & 0.523 & million illegal immigrants & 0.295 \\
    sure joe biden  &  0.659  &  health insurance affordable & 0.206 & poll sleepy joe & 0.396 & trillions new taxes\&amnesty &  0.460 \\
    endorse joe biden &  0.740  & save big health  & 0.199 & live american dream &  0.381 & biden embraced policy & 0.575 \\
    joe biden president & 0.689  & defeat trump gop & 0.608 &  forgotten men woman& 0.185 & policy far left & 0.466 \\
    kamala democrat country & 0.407  &  condemn donald trump  & 0.446 & equal opportunity justice & 0.206 & reduction police funding & 0.231 \\
    \bottomrule
    \end{tabular} 
    \end{adjustbox}
\caption{Most frequent trigrams and their Pearson correlation coefficient with ads stance. P\_Cor: Pearson correlation coefficient. P\_Cor ranges from $−1$ to $1$, with a value of $-1$ meaning a total negative linear correlation, $0$ being no correlation, and $+1$ meaning a total positive correlation.}
\label{tab:triz}
\end{table*}
\begin{figure*}[htbp]
    \begin{subfigure}{1\columnwidth}
  \centering
  \includegraphics[width=\textwidth]{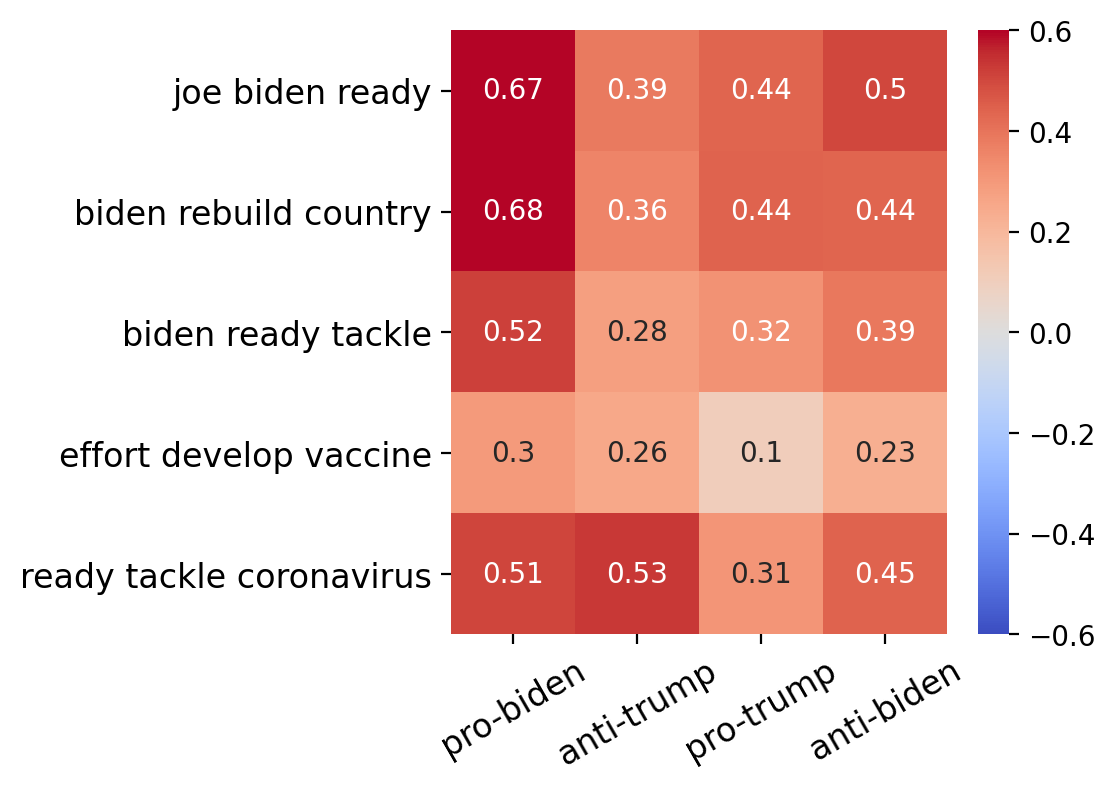}
  \caption{trigrams from pro-biden covid ads}
  \label{fig:covid_pb}
\end{subfigure}%
\begin{subfigure}{1\columnwidth}
  \centering
  \includegraphics[width=\textwidth]{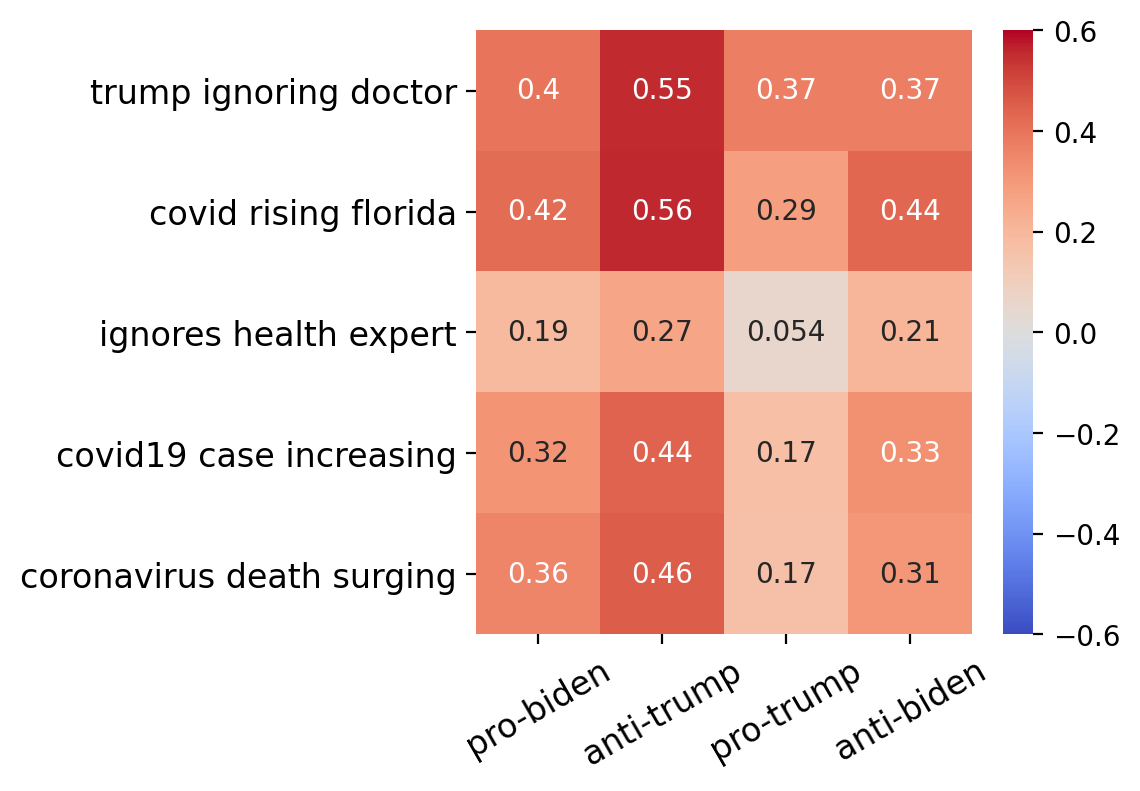}
  \caption{trigrams from anti-trump covid ads}
  \label{fig:covid_at}
\end{subfigure}
\begin{subfigure}{1\columnwidth}
  \centering
  \includegraphics[width=\textwidth]{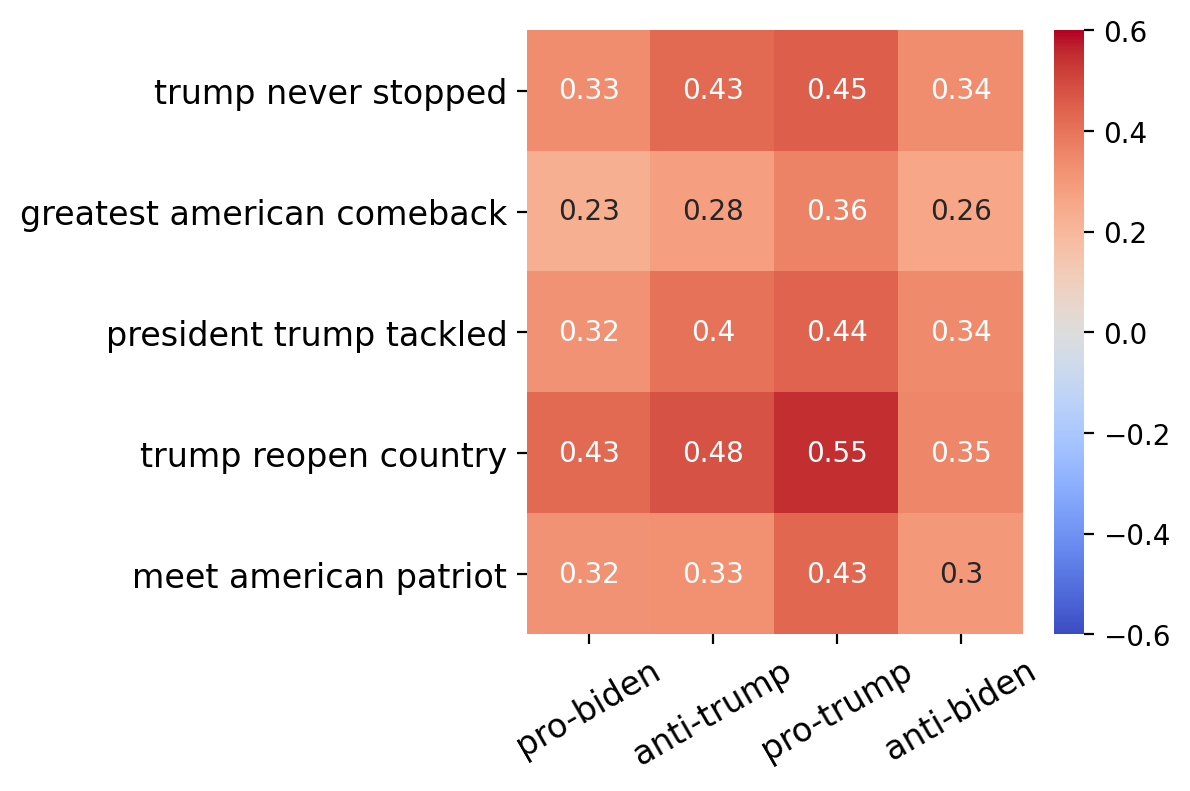}
  \caption{trigrams from pro-trump covid ads}
  \label{fig:covid_pt}
\end{subfigure}%
\begin{subfigure}{1\columnwidth}
  \centering
  \includegraphics[width=\textwidth]{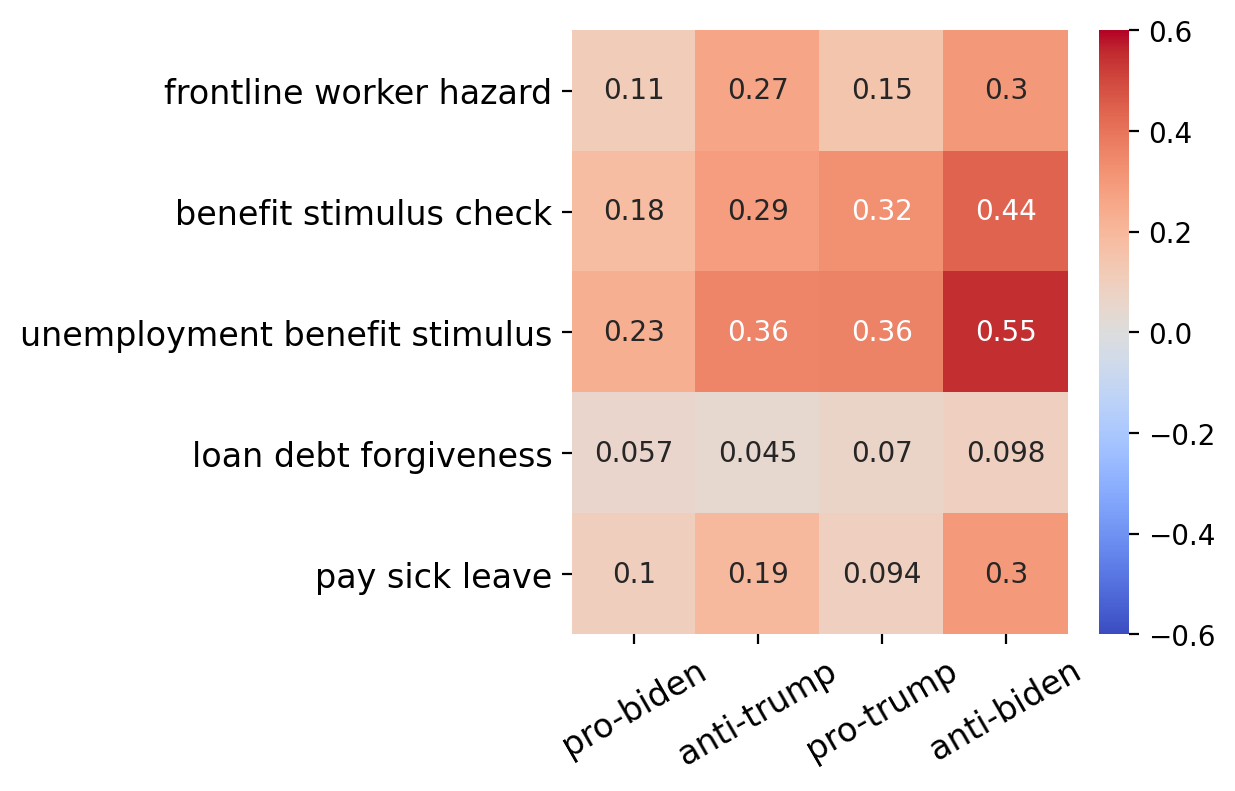}
  \caption{trigrams from anti-biden covid ads}
  \label{fig:covid_ab}
\end{subfigure}
    \caption{Correlation heatmaps for issue-specific (coronavirus) most frequent trigrams and ads stances. The heatmap cell colors represent the percentage of times trigrams appear in the same context.}
    \label{fig:heatmap}
\end{figure*}
\subsection{Qualitative Analysis}
As we do not have ground truths for implicit funding entities, for qualitative evaluation, we consider $50$ funding entities (containing $23881$ ads) that \textbf{do not} explicitly mention any candidate name/party affiliation in their name. To understand their stance as ground truth, we look for their views at \url{OpenSecrets.org}. 
We report results from $25$ liberal and $25$ conservative advertisers.
For each entity, we take the majority vote over the predictions. In Table. \ref{tab:result}, we show the predicted stance with accuracy using our model and compare with ground truth.
\newline
From Table. \ref{tab:result}, we observe \textit{`liberal'} view for the conservative advertiser \textit{`Championing America at Her Best'}\footnote[5]{\url{https://www.championingamerica.com/}}.
The goal of this funding entity is to prevent the reelection of Donald Trump and founding director of this funding entity is Matthew Mattern who is a Republican committed to the ideals of the party of Lincoln and Reagan. Our model predict anti-trump stance for all ads ($100\%$ accuracy) sponsored by this advertiser in our data. 
\newline
We notice that our model provides \textit{`liberal'} view for conservative advertisers \textit{`AMERICANS FOR PROSPERITY'} and \textit{`C3 PAC'} based on majority vote over the predictions (Table. \ref{tab:result}) by predicting `anti-trump' stance mostly. For advertiser \textit{`AMERICANS FOR PROSPERITY'}, we observe that our model predict anti-trump stance where ads contain text like -- \textit{``Make your voice heard today by sending a letter to tell Senator $Y$ – I support the confirmation of Justice Amy Coney Barrett!"}, where $Y$ = any Republican senators. 
Our model predicts anti-trump stance for ads sponsored by \textit{`C3 PAC'} where it contains criticism of Trump inside quote, i.e., \textit{We can't let Nancy Pelosi get away with more baseless lies and sham conspiracy theories. Join us in condemning her recent ``Trump and GOP lawmakers are Enemies of the People" remarks.}
Though our model can predict \textit{`liberal'} views for for all liberal advertisers based on majority vote (Table. \ref{tab:result}), we notice that our model predicts pro-trump stance for the ads having rhetorical question -- \textit{``Joe Biden worked with Obama to save the auto industry and bring back jobs. What has Trump done for American workers?"}
\subsection{Descriptive Insights}
\label{Descriptive_Insights}
In our data, we have more \textit{pro-trump} stance ads. We notice that \textit{supreme court} is the most and \textit{lgbtq} is the least prominent issues in the ad content.
%
To answer \textbf{RQ2}, we focus on how distinctive the contents of `$X-trump$' and `$X-biden$' ads, where $X = pro, anti$. 
To understand what kind of words ads use to represent positive and negative stance
towards the candidates, we show most frequent trigrams for each category (Table. \ref{tab:triz}). 
To compute the statistical significance, we calculate Pearson correlation coefficient \cite{pearson1895notes} between ads of each stance category and each trigram generated by corresponding category of ads ($pearson\_corr(t_i,s_j)$, where $t_i$ = trigrams generated from $s_j$ and $s_j$ = ads related to each stance category). To calculate correlation, we provide the embedding for each trigrams and embedding of ads for each stance category. For embedding, we use
Sentence-BERT embeddings \cite{reimers2019sentence}.
Trigrams and corresponding Pearson correlation coefficient are shown in Table. \ref{tab:triz}. 
From Table. \ref{tab:triz}, we notice that the ads having pro-biden stance has common trigrams like `vote joe biden', `today vote democrat', `joe kamala democrat' etc, whether anti-trump has trigrams like `defeat donald trump', `request ballot today', `new trumpcare plan' etc. On the other hand, ads with pro-trump stance has common trigrams like `president trump need', `vote november $3^{rd}$', `live american dream', `equal opportunity justice' etc, whether anti-biden ads contains trigrams like `radical left taken', `million illegal immigrants', `trillions new taxes\&amnesty', `reduction police funding' etc. 
Biden's economy \& tax plans and immigration policies are often the target of anti-biden attacks, and as such are often mentioned and
accused of \textit{raising tax on middle class families},  \textit{providing illegal immigrants amnesty and healthcare}, \textit{embracing far left policy}. On the other side, pro-trump
messages focus on \textit{ Trump's vow never to forget the `forgotten men and women'}. They also advertise about \textit{how every single citizen have a chance to live their American dream and have equal opportunity justice} under Trump administration. 

To show the noticeable qualitative differences, we create wordcloud with the most frequent words from the whole data. Fig. \ref{fig:stance_wc} shows the visual representation of the text for each stance category. Words are usually single words, and the importance of each tag is shown with font size and color.
%
Ads with pro-biden and anti-trump stances are talking about \textit{vote, chip, donation, win, defeat, elect, ballot, win, support} (Fig. \ref{fig:pro_biden} and \ref{fig:anti_trump}). Positive stance regarding Trump has following words \textit{president, vote, fighting, fake news, real american}  (Fig. \ref{fig:pro_trump}). Anti-biden wordcloud has noticeable words such as \textit{radical left, sleepy, washington swamp, illegal immigrant, far left} (Fig. \ref{fig:anti_biden}). 


\subsubsection{Issue-specific Ads}
\label{Ads_Issue}
To analyze \textit{how do ads talk about an issue based on stances}, i.e., pro trump vs. anti-biden ads on immigration e.g., ``Trump will build the wall" vs. ``Biden is weak on immigration",
we put a condition on meaningful words that characterize an issue. Then, we calculate the correlation between ads and the most frequent trigrams in each categorty ($pearson\_corr(t_{i,k},s_{j,k})$, where $t_{i,k}$ = trigrams generated from $s_j$ on $k$ issue and $s_j$ = ads related to each stance category on $k$ issue). 
To better understand how each stance represents messaging, we analyze their co-occurrence and convey this information in heatmaps. Each row in the heatmap captures the association strength between issue-specific trigrams ($y$-axis) and all the stances ($x$-axis). The heatmap cell colors represent the percentage of times trigrams appear in the same context.
The heatmaps showing the correlation of trigrams and stances on `coronavirus' issue are shown in Fig. \ref{fig:heatmap}. It demonstrates how trigrams are correlated to each stance category.
%
Fig. \ref{fig:covid_pb} shows the top $5$ most frequent trigrams from pro-biden ads focusing on the `coronavirus' issue and their Pearson correlation coefficients for each stance category. For example, there is a higher correlation between trigrams that focuses on \textit{Biden's plan for tackling coronavirus by putting effort to develop vaccine, rebuilding country} and pro-biden stance. On the other hand, anti-trump ads correlate more with the trigrams focusing on \textit{rising covid cases and deaths in Florida and Trump's ignorance regarding health experts} (Fig. \ref{fig:covid_at}). \textit{Trump's plan for reopening country} is one of the prominent messages of pro-trump ads (Fig. \ref{fig:covid_pt}). From Fig. \ref{fig:covid_ab} we notice that anti-biden ads highly correlate with \textit{coronavirus stimulus}. 
A more thorough qualitative examination of the content of these ads is left for future work, and instead we focus on their audience reach.

\subsection{Audience Demographics}
\label{Audience_Demographic}
In this section, we focus on the audience of these ad campaigns. As Facebook Ads Library provides summary of demographic statistics on `impressions' received by each ad, as a distribution over 3 genders (male, female, unknown) and 7 age groups. This metric describes the views of each ad, which may be different from users exposed to the ad, as the same user may be exposed multiple times. To answer \textbf{RQ3}, we analyze 1) Targeted demographics by the advertisers, 2) Ad impressions by the demographics.


\begin{figure*}
	\centering
	\begin{subfigure}{.5\textwidth}
		\includegraphics[width=\textwidth]{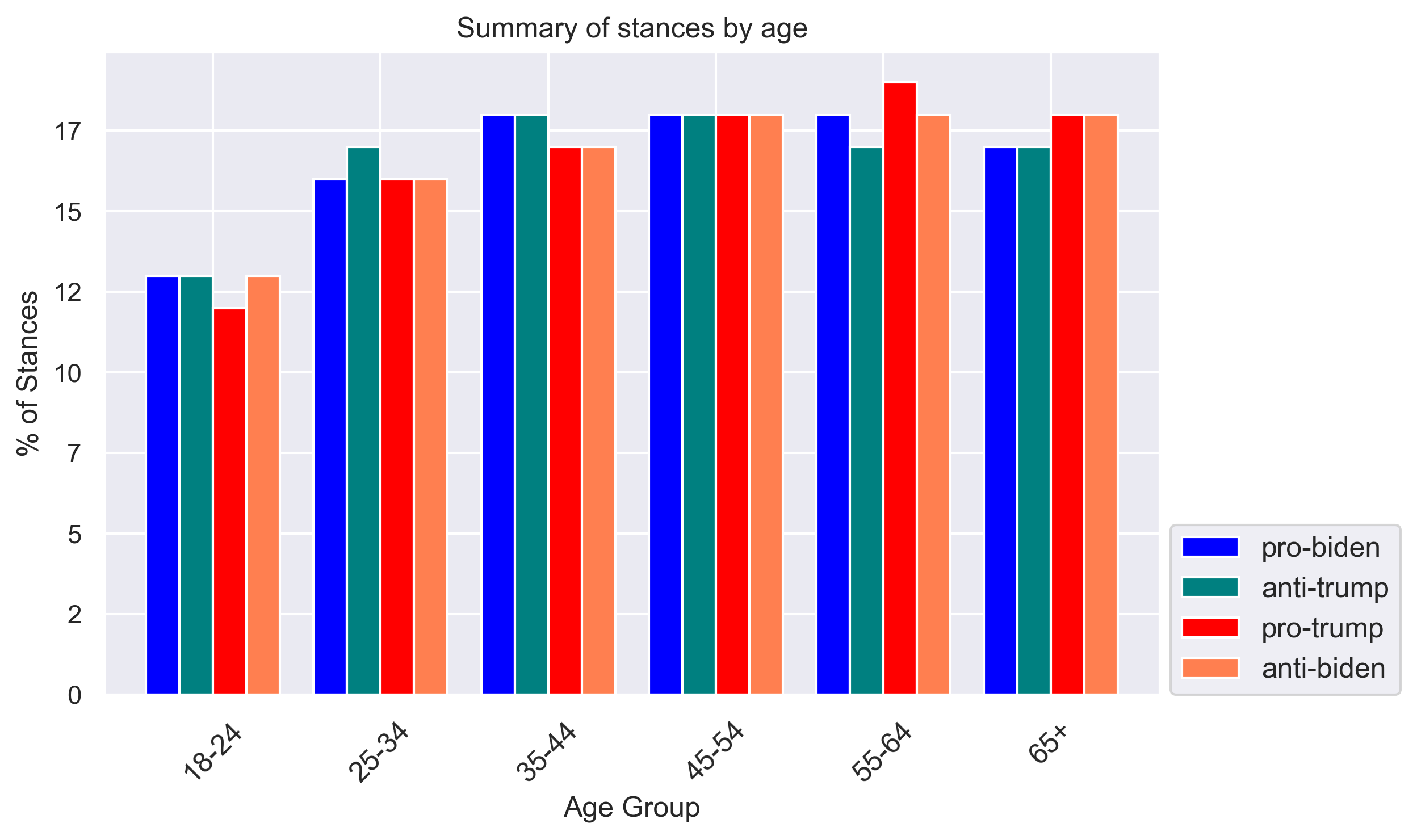}
		\caption{}
		\label{fig:stance_age_per}
	\end{subfigure}%
	\begin{subfigure}{.5\textwidth}
		\includegraphics[width=\textwidth]{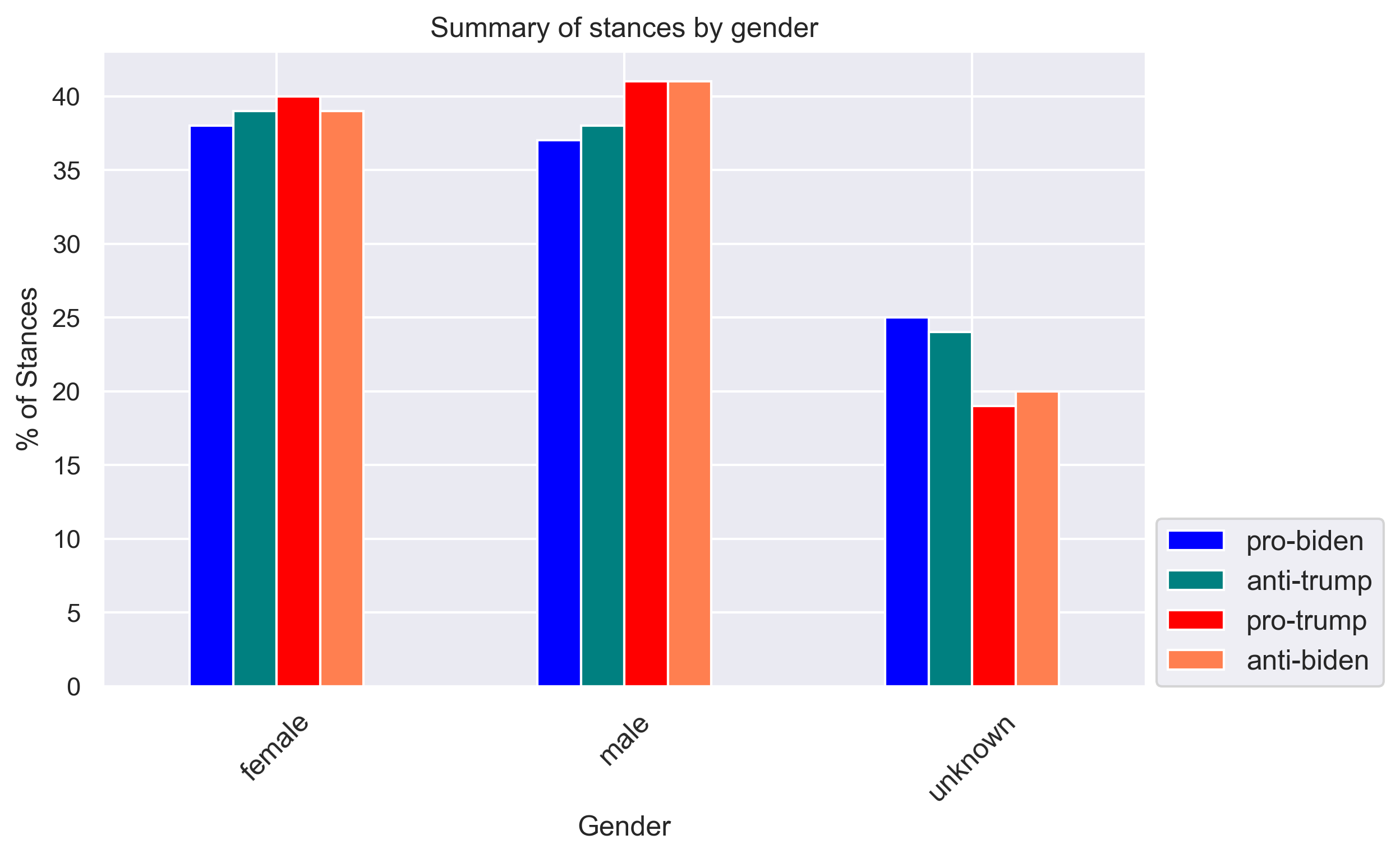}
		\caption{}
		\label{fig:stance_gen_per}
	\end{subfigure}
	\caption{demographic distribution (\textbf{a.} age and \textbf{b.} gender) over percentages of ads of each stance category. Chi-square test results indicate an association between audience's demographics and advertisers' stances (p-value $< 0.05$). Pro-biden and anti-trump ads mostly target $35-54$ age group and the female gender. Advertisers with pro-trump and anti-biden stances target the $55-64$ age group and the male population mostly. The age group $45-54$ is equally targeted by all four stances.}
    \label{fig:stance_demo_per}
\end{figure*}

\begin{table*}
\begin{adjustbox}{width=1\textwidth,center}
\begin{tabular}{llcc}
    \toprule
    \textbf{Null Hypothesis,} \boldmath{$H_0$} & \textbf{Alternate Hypothesis,} \boldmath{$H_a$} & \textbf{T-test statistics} & \textbf{P-value}   \\
     \midrule
     More females than males of all ages & More females than males from all ages & $4.768^{**}$ & $0.005$ \\
     do not watch pro-biden ads.  &  watch pro-biden ads.  \\

     More females than males of all ages & More females than males from all ages & $4.847^{**}$ & $0.004$ \\ do not watch anti-trump ads.  & watch anti-trump ads. \\
      More males than females from age &  More males than females from age  &  $4.841^*$ & $0.017$ \\ range 18-54 do not view pro-trump ads.  & range 18-54 view pro-trump ads. \\

        More males than females from age &  More males than females from age  &  $4.131^*$ & $0.026$  \\ range 18-54 do not view anti-biden ads. & range 18-54 view anti-biden ads. \\ 

        More females than males from older & More females than males from older  &  $1.255^{ns}$ & $0.428$ \\ age (55+) do not watch pro-trump ads.  &   age (55+) watch pro-trump ads.    \\

        More females than males from older &  More females than males from older  &  $1.057^{ns}$ & $0.483$  \\ age (55+) do not watch anti-biden ads. & age (55+) watch anti-biden ads. \\
        
      
    \bottomrule
    \end{tabular} 
\end{adjustbox}
\caption{T-test of the influence of audiences' age and gender on ad impressions. ** = highly statistically significant at p-value $< 0.01$; * = statistically significant at p-value $< 0.05$; ns = statistically not significant (p-value $> 0.05$). }
\label{tab:t_test}
\end{table*}
\begin{figure*}[htbp]
  \centering  
  \includegraphics[width= \textwidth]{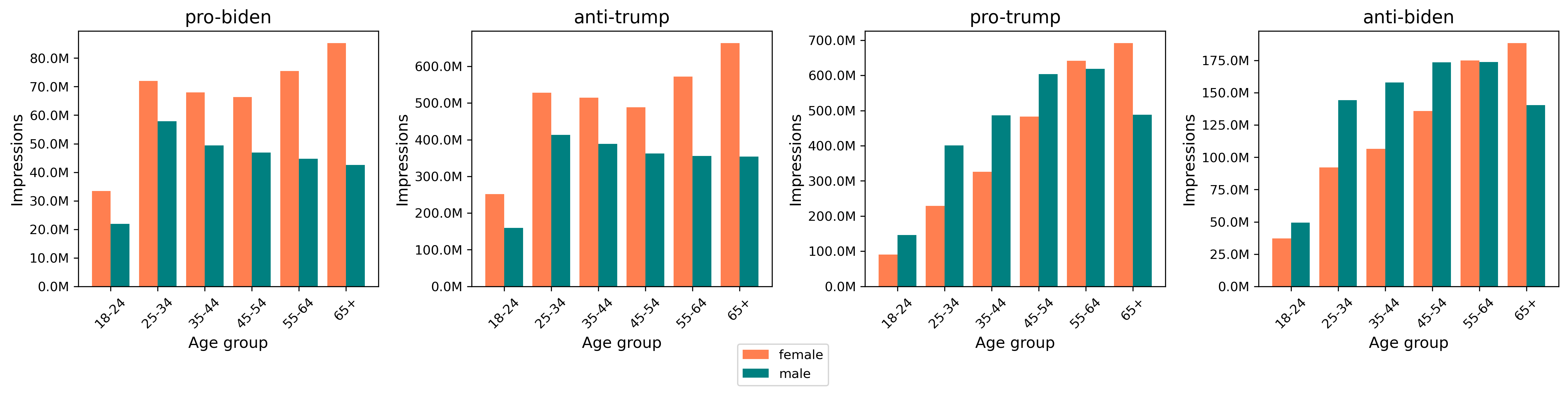}
    \caption{Distribution of impressions over age and gender in each stance category. Pro-biden and anti-trump ads are viewed mainly by female audiences and it's highly statistically significant (p-value $< 0.01$). Pro-trump and anti-biden ads are mostly viewed by males from age range $18-54$ (p-value $< 0.05$). Results of t-test hypothesis testing are shown in Table \ref{tab:t_test}.}
    \label{fig:agi}
\end{figure*}
\subsubsection{Targeted Demographics}
We have $3.4\%$ pro-biden ads, $55\%$ pro-trump, $13.4\%$ anti-biden, and $28.2\%$ anti-trump ads targeting different demographics. To answer the question ``Is there any association between audience's demographics and advertisers' stances?", we perform chi-square test of contingency \cite{cochran1952chi2}. The chi-square test provides a method for testing the association between the row and column variables in a two-way table called contingency table. The null hypothesis $H_0$ assumes that there is no association between the variables, while the alternative hypothesis $H_a$ claims that some association does exist. The chi-square test statistic is computed as $\chi^2 = \sum \frac{(observed - expected)^2 } {expected} $.
The distribution of the statistic $\chi^2$ is denoted as $\chi^{2}_{(df)}$, where $df$ is the number of degrees of freedom. $df = (r-1)(c-1)$, where $r$ represents the number of rows and $c$ represents the number of columns in the contingency table. The p-value for the chi-square test is the probability of observing a value at least as extreme as the test statistic for a chi-square distribution with $(r-1)(c-1)$ degrees of freedom.
%
To perform a chi-square test, we take age distribution over stance and gender distribution over stance separately to build contingency tables correspondingly. We choose value of significance level, $\alpha = 0.05$. The p-value for both cases is $< 0.05$, which is statistically significant and we reject the null hypothesis $H_0$, indicating that there is some association between the audience's demographics and advertisers' stances. 
Fig. \ref{fig:stance_demo_per} shows the comparison of the percentages of stances by age group (age group $13-17$ has been dropped due to no activity) (Fig. \ref{fig:stance_age_per}) and gender (Fig. \ref{fig:stance_gen_per}). We notice that the ads with pro-biden and anti-trump stances mostly target the age group $35-54$, whereas pro-trump ads mainly target age rage $55-64$ and anti-trump ads focus $45-65+$ age group. Interestingly age group $45-54$ is equally targeted by `$X-trump$' and `$X-biden$' ads, where $X = pro, anti$ (Fig. \ref{fig:stance_age_per}). However, ads with pro-biden and anti-trump stances target the female population mostly, while pro-trump and anti-biden ads target the male population mostly (Fig. \ref{fig:stance_gen_per}). We can conclude that 
funding agencies supporting Biden in election targeting the age group $35-54$ and women and funding agencies supporting Trump in election targeting older people and men are not due to random variation.

\subsubsection{Demographic Ad Impression}
Among the $12.5B$ impressions received by the ads we have collected, ads with pro-biden stance has $5.3\%$ impressions, $41.6\%$ for pro-trump, anti-biden $12.6\%$, anti-trump has $40.4\%$ impressions. Fig. \ref{fig:agi} shows the  distribution of impressions over demographics (age group $13-17$ and unknown gender have been dropped because of having almost no activity). 
Considering all ads, pro-trump advertising has the majority of impressions and it is mostly viewed by younger male than female audiences. Both pro-trump and anti-biden ads have more views from males than females in age range $18-54$.
Fig. \ref{fig:agi} shows that more women from all age groups, compared to men from all age groups, watch pro-biden and anti-trump ads. We note that the female audience of ads is even more skewed towards older age than males.
We perform t-test \cite{student1908probable} hypothesis testing to provide statistical evidence of our study. Table \ref{tab:t_test} shows the null hypothesis ($H_0$), alternate hypothesis ($H_a$), t-test statistics with p-value for each tested variables. In the t-test, our level of significance, $\alpha = 0.05$. If $ p-value > \alpha$, we accept $H_0$; otherwise we reject $H_0$. No statistical significance is found when we test whether more females compared to males from older age watch pro-trump and anti-biden ads as p-value $> 0.05$ and we accept $H_0$ (last $2$ rows of Table \ref{tab:t_test}). Results from top $4$ rows in Table \ref{tab:t_test} show that pro-biden and anti-trump ads are viewed mostly by females than males regardless of age group, whereas more males than females from age group $18-54$ tend to watch pro-trump and anti-biden ads (p-value $< 0.05$ and reject $H_0$).

\begin{figure*}
\begin{subfigure}{1\textwidth}
  \centering  
  \includegraphics[width= 1\textwidth]{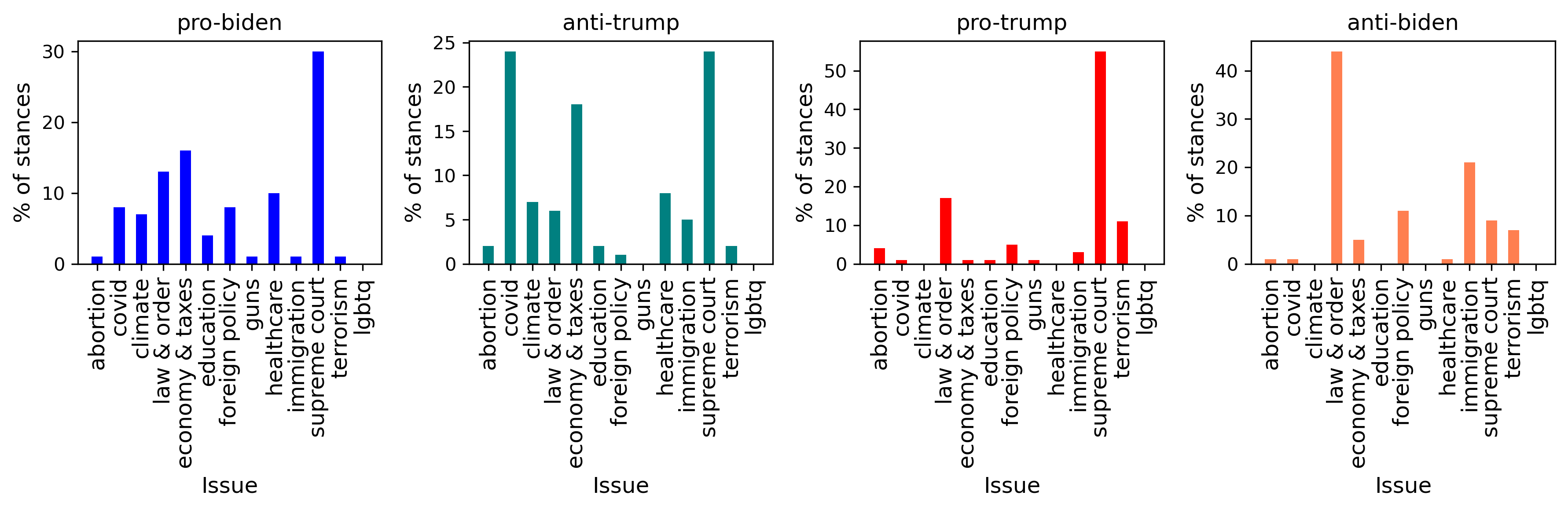}
    \caption{Pennsylvania}
    \label{fig:pa_issue_stance_dis}
\end{subfigure}
\begin{subfigure}{1\textwidth}
  \centering  
  \includegraphics[width= 1\textwidth]{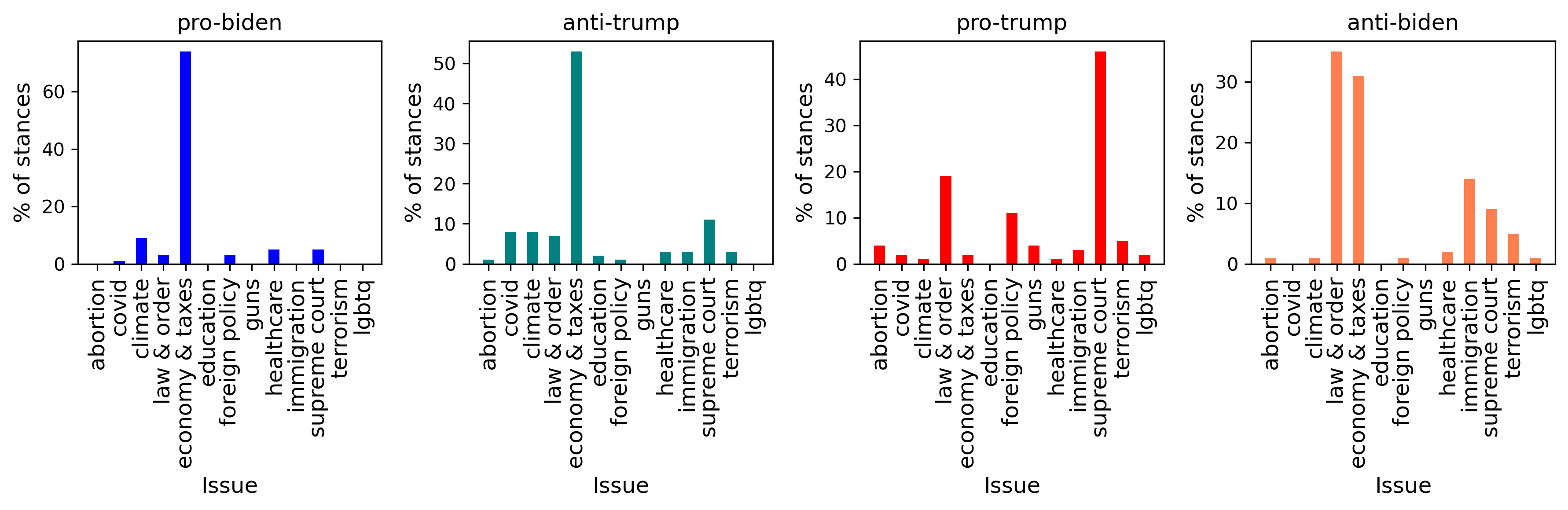}
    \caption{New York}
    \label{fig:ny_issue_stance_dis}
\end{subfigure}
\begin{subfigure}{1\textwidth}
  \centering
  \includegraphics[width= 1\textwidth]{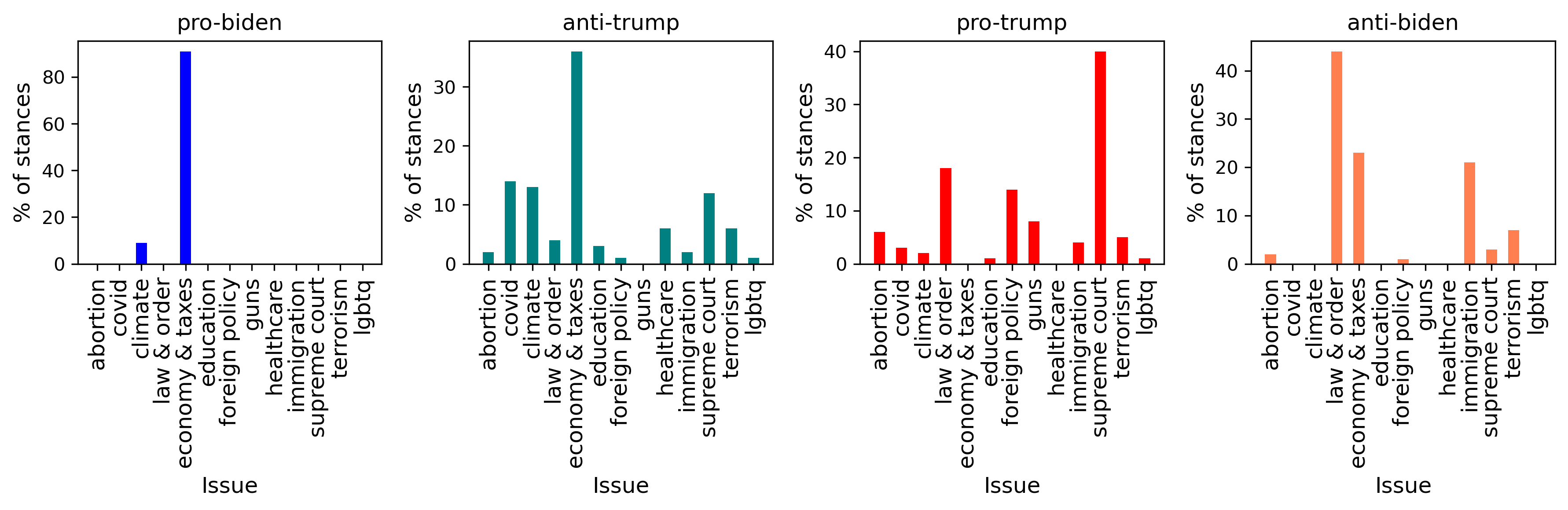}
    \caption{Idaho}
    \label{fig:id_issue_stance_dis}
\end{subfigure}
\caption{Issue distribution over percentages of ads of each stance category in (a) PA, (b) NY, (c) ID considering ads having $> 10\%$ regional impressions. Chi-square test results indicate an association between advertisers' issue and stances focusing on a specific region (p-value $< 0.05$). Pro-biden and anti-trump ads focus on various issues, e.g., `covid', `economy \& taxes', `supreme court', depending on region. Conversely, pro-trump and anti-biden stances are noticeable in `supreme court' and `law \& order' related ads, respectively, targeting all three states.}

\label{fig:3states}
\end{figure*}

\begin{figure}[htbp]
  \centering  
  \includegraphics[width= 1\columnwidth]{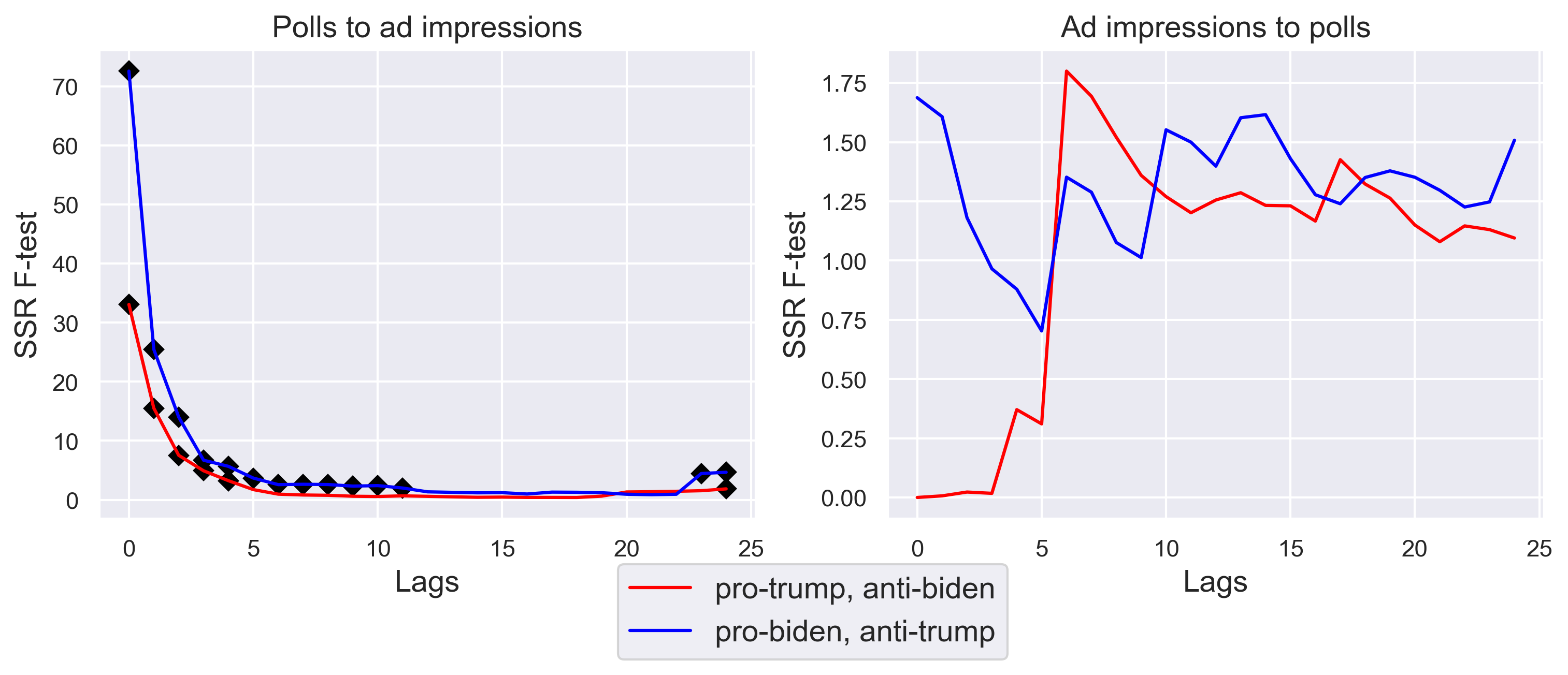}
    \caption{Granger causality tests comparing polling averages for each day and  ad impressions. On the $x$-axis, we report the number of lags in days we consider for the delay between the two time series; on the $y$-axis, we show the sum-of-square $F-test$ statistics for corresponding lag. We highlight those with p-value $< 0.05$ with black diamond box.}
    \label{fig:gc}
\end{figure}
\subsection{State-wise Issue and Demographics}
\label{Geographic}
We investigate state-wise issue and demographic distribution over stances to answer \textbf{RQ4}. We use historical voting data\footnote[6]{\url{https://www.visualcapitalist.com/u-s-presidential-voting-history-by-state/}} to determine red/blue/battleground status of the states. 
We consider battleground state Pennsylvania ($451576$ ads), blue state New York ($389788$ ads), and red state Idaho ($306476$ ads) to analyze their stances, issues, and corresponding demographics. To narrow down our analysis, we consider ads having $> 10\%$ regional impressions in these states.
We perform chi-square test by taking issue distribution over stance to build contingency tables separately for PA, NY, and ID. We choose the value of significance level, $\alpha = 0.05$, and our test results show that the p-value $< 0.05$ for all three states. Therefore, we reject the null hypothesis, indicating some association between advertisers' issues and stances focusing on the region. 
Fig. \ref{fig:pa_issue_stance_dis} shows issue distribution over stances in the swing state PA. In PA, we notice the highest pro-biden stance on `supreme court' issue, and a higher anti-trump issue on `covid' and `supreme court'.  
We show issue distribution with stances in the blue state NY and red state ID in Fig. \ref{fig:ny_issue_stance_dis} and Fig. \ref{fig:id_issue_stance_dis} correspondingly. Issue like `economy \& taxes' has more pro-biden and anti-trump stances both in NY and ID (Fig. \ref{fig:id_issue_stance_dis}). On the other hand, anti-biden ads mostly focus on `law \& order' and `economy \& taxes' issues in New York (Fig. \ref{fig:ny_issue_stance_dis}).
It's noticeable that `supreme court’ is the most prominent issue in pro-trump ads in PA, NY, and ID. In addition, anti-biden stances are apparent in ads related to the `criminal justice reform, race, law \& order' issue for all three states.

%
To analyze age and gender distribution with stances based on the three states, we again perform chi-square test by taking age and gender distribution over stance to build contingency tables separately. Our test results show the p-value $< 0.05$ for age and gender in PA, NY, and ID. Therefore, we reject the null hypothesis, indicating some association between advertisers' stances and demographics focusing on the region.
In PA, pro-biden and anti-trump ads primarily target the younger population ($25-44$). In contrast, pro-trump and anti-biden ads mainly target the people of $45-65+$ in PA. We notice that ads with pro-trump and anti-biden stances target the age group $55-64$ in NY. Interestingly in ID, pro-biden ads are more prominent, and the target age group is $55-64$. 
After analyzing gender distribution with stances, we notice that more men than women are targeted by pro-trump and anti-biden ads in all three states. In PA, both females and males are targeted equally by pro-biden and anti-trump ads. More females compared to males are targeted by anti-trump ads in NY. On the other hand, more female populations (60\%) than males are targeted by pro-biden ads in Idaho.

\subsection{Granger Causality with Polls}
\label{Granger}
To start off, toward answering \textbf{RQ5}, we collect poll data of year $2020$ from \url{https://fivethirtyeight.com/}.
We exclude from the analysis the period after $3^{rd}$ November 2020.
We compute two time series:
\newline
(1) For poll data, we take the sum of average poll count of each presidential candidate for each day called $Polls(t)$.
\newline
(2) For ads, we calculate the total number of impressions of `$X-trump$' and `$X-biden$' ads, where $X = pro, anti$ for a given day, $AdImpressions(t)$.
\newline
We compute the following two Granger causality tests with these two time series to check (1) Does $Polls(t)$ \textit{Granger cause} $AdImpressions(t)$? (2) Does $AdImpressions(t)$ \textit{Granger cause} $Polls(t)$? The null hypothesis $(H_0)$ assumed by the first test is that $Polls(t)$ does not Granger cause $AdImpressions(t)$ and the alternative hypothesis $(H_A)$ is that $Polls(t)$ Granger causes $AdImpressions(t)$. For the second test, $H_0$ is $AdImpressions(t)$ does not Granger cause $Polls(t)$ and $H_A$ = $AdImpressions(t)$ Granger causes $Polls(t)$. We reject $H_0$ if p-value is $< 0.05$ for these tests.
We report results for these two tests in Fig. \ref{fig:gc}. We notice a significant F-test for the hypothesis of $Polls(t)$ Granger causes $AdImpressions(t)$ for ads with both stances (Left side of Fig. \ref{fig:gc}). Conversely, we find no significant Granger causality from ad impressions to polls (Right side of Fig. \ref{fig:gc}). Our finding from this analysis is when polls lean towards one candidate, Facebook ads sponsored by the advertisers supporting that candidate get more attention.

\section{Conclusion}
We suggest a weakly supervised approach for analyzing political campaigns on social media, and show the difference in the frequency of positive and negative ads and their main policy issue, as a function of the group targeted. By answering our research questions and providing statistical tests to discern the most significant findings, we inform empirical understandings of how polarized political environments are linked to the underlying funding structure for election ads. Though our approach is tailored for the specific use case, this can be extended to analyze any kind of ads on social media. 

\section{Ethical Impact}
The data collected in the paper was made publicly available by Facebook Ads API and does not contain any personal information. Any qualitative result that we report is an outcome from a machine learning model that does not represent the authors’ personal views.

\section*{Acknowledgements}
We are thankful to the anonymous reviewers for their insightful comments. This work was partially supported by Purdue Graduate School Summer Research Grant (to TI) and an NSF CAREER award IIS-2048001.

\bibliography{aaai23}

\begin{thebibliography}{60}
\providecommand{\natexlab}[1]{#1}

\bibitem[{Andreou et~al.(2019)}]{andreou2019measuring}
Andreou, A.; et~al. 2019.
\newblock Measuring the Facebook advertising ecosystem.
\newblock In \emph{NDSS}.

\bibitem[{Badawy, Ferrara, and Lerman(2018)}]{8508646}
Badawy, A.; Ferrara, E.; and Lerman, K. 2018.
\newblock Analyzing the Digital Traces of Political Manipulation: The 2016
  Russian Interference Twitter Campaign.
\newblock In \emph{ASONAM}.

\bibitem[{Belkin, Niyogi, and Sindhwani(2006)}]{belkin2006manifold}
Belkin, M.; Niyogi, P.; and Sindhwani, V. 2006.
\newblock Manifold regularization: A geometric framework for learning from
  labeled and unlabeled examples.
\newblock \emph{JMLR}.

\bibitem[{Capozzi et~al.(2020)}]{capozzi2020facebook}
Capozzi, A.; et~al. 2020.
\newblock Facebook Ads: Politics of Migration in Italy.
\newblock In \emph{ICSI}.

\bibitem[{Capozzi et~al.(2021)}]{capozzi2021clandestino}
Capozzi, A.; et~al. 2021.
\newblock Clandestino or Rifugiato? Anti-immigration Facebook Ad Targeting in
  Italy.
\newblock In \emph{CHI}.

\bibitem[{Church and Hanks(1990)}]{church1990word}
Church, K.~W.; and Hanks, P. 1990.
\newblock Word association norms, mutual information, and lexicography.
\newblock \emph{Computational linguistics}.

\bibitem[{Cochran(1952)}]{cochran1952chi2}
Cochran, W.~G. 1952.
\newblock The $\chi$2 test of goodness of fit.
\newblock \emph{The Annals of mathematical statistics}.

\bibitem[{Cohen(1960)}]{cohen1960coefficient}
Cohen, J. 1960.
\newblock A coefficient of agreement for nominal scales.
\newblock \emph{EPM}.

\bibitem[{Devlin et~al.(2019)}]{devlin2019bert}
Devlin, J.; et~al. 2019.
\newblock BERT: Pre-training of Deep Bidirectional Transformers for Language
  Understanding.
\newblock In \emph{HLT-NAACL}.

\bibitem[{Durant and Smith(2006)}]{durant2006predicting}
Durant, K.~T.; and Smith, M.~D. 2006.
\newblock Predicting the political sentiment of web log posts using supervised
  machine learning techniques coupled with feature selection.
\newblock In \emph{KDWEB}.

\bibitem[{Ferrara et~al.(2020)}]{ferrara2020characterizing}
Ferrara, E.; et~al. 2020.
\newblock Characterizing social media manipulation in the 2020 US presidential
  election.
\newblock \emph{First Monday}.

\bibitem[{Field, Kliger et~al.(2018)}]{field2018framing}
Field, A.; Kliger, D.; et~al. 2018.
\newblock Framing and Agenda-setting in Russian News: a Computational Analysis
  of Intricate Political Strategies.
\newblock In \emph{EMNLP}.

\bibitem[{Field and Tsvetkov(2019)}]{field2019entity}
Field, A.; and Tsvetkov, Y. 2019.
\newblock Entity-Centric Contextual Affective Analysis.
\newblock In \emph{ACL}.

\bibitem[{Giorgi et~al.(2020)}]{giorgi2020declutr}
Giorgi, J.~M.; et~al. 2020.
\newblock Declutr: Deep contrastive learning for unsupervised textual
  representations.
\newblock \emph{arXiv:2006.03659}.

\bibitem[{Granger(1988)}]{granger1988some}
Granger, C.~W. 1988.
\newblock Some recent development in a concept of causality.
\newblock \emph{Journal of econometrics}.

\bibitem[{Greene and Resnik(2009)}]{greene2009more}
Greene, S.; and Resnik, P. 2009.
\newblock More than words: Syntactic packaging and implicit sentiment.
\newblock In \emph{HLT-NAACL}.

\bibitem[{Grover and Leskovec(2016)}]{grover2016node2vec}
Grover, A.; and Leskovec, J. 2016.
\newblock node2vec: Scalable Feature Learning for Networks.
\newblock In \emph{KDD}.

\bibitem[{Han and Shen(2016)}]{han2016partially}
Han, Y.; and Shen, Y. 2016.
\newblock Partially Supervised Graph Embedding for Positive Unlabelled Feature
  Selection.
\newblock In \emph{IJCAI}.

\bibitem[{Hersh(2015)}]{hersh2015}
Hersh, E.~D. 2015.
\newblock \emph{Hacking the electorate: How campaigns perceive voters}.
\newblock Cambridge University Press.

\bibitem[{Hisano(2018)}]{hisano2018semi}
Hisano, R. 2018.
\newblock Semi-supervised graph embedding approach to dynamic link prediction.
\newblock In \emph{CompleNet}.

\bibitem[{Hochreiter and Schmidhuber(1997)}]{hochreiter1997long}
Hochreiter, S.; and Schmidhuber, J. 1997.
\newblock Long short-term memory.
\newblock \emph{Neural computation}.

\bibitem[{Islam and Goldwasser(2022{\natexlab{a}})}]{islam2021twitter}
Islam, T.; and Goldwasser, D. 2022{\natexlab{a}}.
\newblock Twitter User Representation Using Weakly Supervised Graph Embedding.
\newblock In \emph{ICWSM}.

\bibitem[{Islam and Goldwasser(2022{\natexlab{b}})}]{islam2022covidfbAd}
Islam, T.; and Goldwasser, D. 2022{\natexlab{b}}.
\newblock Understanding COVID-19 Vaccine Campaign on Facebook using Minimal
  Supervision.
\newblock In \emph{IEEE Big Data}.

\bibitem[{Iyyer et~al.(2014)}]{iyyer2014political}
Iyyer, M.; et~al. 2014.
\newblock Political ideology detection using recursive neural networks.
\newblock In \emph{ACL}.

\bibitem[{Jensen(2017)}]{jensen2017social}
Jensen, M.~J. 2017.
\newblock Social media and political campaigning: Changing terms of engagement?
\newblock \emph{IJPP}.

\bibitem[{Johnson and Goldwasser(2016)}]{johnson2016all}
Johnson, K.; and Goldwasser, D. 2016.
\newblock “All I know about politics is what I read in Twitter”: Weakly
  Supervised Models for Extracting Politicians’ Stances From Twitter.
\newblock In \emph{COLING}.

\bibitem[{Kingma and Ba(2014)}]{kingma2014adam}
Kingma, D.; and Ba, J. 2014.
\newblock Adam: A method for stochastic optimization.
\newblock \emph{arXiv:1412.6980}.

\bibitem[{Klebanov, Beigman, and Diermeier(2010)}]{klebanov2010vocabulary}
Klebanov, B.~B.; Beigman, E.; and Diermeier, D. 2010.
\newblock Vocabulary choice as an indicator of perspective.
\newblock In \emph{ACL}.

\bibitem[{Kushin and Yamamoto(2010)}]{kushin2010did}
Kushin, M.~J.; and Yamamoto, M. 2010.
\newblock Did social media really matter? College students' use of online media
  and political decision making in the 2008 election.
\newblock \emph{MCS}.

\bibitem[{Lin et~al.(2006)}]{lin2006side}
Lin, W.; et~al. 2006.
\newblock Which side are you on?: identifying perspectives at the document and
  sentence levels.
\newblock In \emph{CoNLL}.

\bibitem[{Loshchilov and Hutter(2018)}]{loshchilov2018decoupled}
Loshchilov, I.; and Hutter, F. 2018.
\newblock Decoupled Weight Decay Regularization.
\newblock In \emph{ICLR}.

\bibitem[{Marozzo and Bessi(2018)}]{marozzo2018analyzing}
Marozzo, F.; and Bessi, A. 2018.
\newblock Analyzing polarization of social media users and news sites during
  political campaigns.
\newblock \emph{SNAM}.

\bibitem[{Meng et~al.(2012)}]{meng2012entity}
Meng, X.; et~al. 2012.
\newblock Entity-centric topic-oriented opinion summarization in twitter.
\newblock In \emph{ACM SIGKDD}.

\bibitem[{Mitchell et~al.(2013)}]{mitchell2013open}
Mitchell, M.; et~al. 2013.
\newblock Open domain targeted sentiment.
\newblock In \emph{EMNLP}.

\bibitem[{Mohammad et~al.(2016)}]{mohammad2016semeval}
Mohammad, S.; et~al. 2016.
\newblock Semeval-2016 task 6: Detecting stance in tweets.
\newblock In \emph{SemEval}.

\bibitem[{Pearson(1895)}]{pearson1895notes}
Pearson, K. 1895.
\newblock Notes on Regression and Inheritance in the Case of Two Parents
  Proceedings of the Royal Society of London, 58, 240-242.

\bibitem[{Pennington, Socher, and Manning(2014)}]{pennington2014glove}
Pennington, J.; Socher, R.; and Manning, C. 2014.
\newblock Glove: Global vectors for word representation.
\newblock In \emph{EMNLP}.

\bibitem[{Perozzi, Al-Rfou, and Skiena(2014)}]{perozzi2014deepwalk}
Perozzi, B.; Al-Rfou, R.; and Skiena, S. 2014.
\newblock Deepwalk: Online learning of social representations.
\newblock In \emph{KDD}.

\bibitem[{Ratkiewicz et~al.(2011)}]{ratkiewicz2011detecting}
Ratkiewicz, J.; et~al. 2011.
\newblock Detecting and tracking political abuse in social media.
\newblock In \emph{ICWSM}.

\bibitem[{Recasens et~al.(2013)}]{recasens2013linguistic}
Recasens, M.; et~al. 2013.
\newblock Linguistic models for analyzing and detecting biased language.
\newblock In \emph{ACL}.

\bibitem[{Reimers and Gurevych(2019)}]{reimers2019sentence}
Reimers, N.; and Gurevych, I. 2019.
\newblock Sentence-bert: Sentence embeddings using siamese bert-networks.
\newblock \emph{arXiv:1908.10084}.

\bibitem[{Ribeiro et~al.(2019)}]{ribeiro2019microtargeting}
Ribeiro, F.~N.; et~al. 2019.
\newblock On microtargeting socially divisive ads: A case study of
  russia-linked ad campaigns on facebook.
\newblock In \emph{ACM FAccT}.

\bibitem[{Roy and Goldwasser(2020)}]{roy2020weakly}
Roy, S.; and Goldwasser, D. 2020.
\newblock Weakly Supervised Learning of Nuanced Frames for Analyzing
  Polarization in News Media.
\newblock In \emph{EMNLP}.

\bibitem[{Schuster and Paliwal(1997)}]{schuster1997bidirectional}
Schuster, M.; and Paliwal, K.~K. 1997.
\newblock Bidirectional recurrent neural networks.
\newblock \emph{IEEE transactions on Signal Processing}.

\bibitem[{Serrano et~al.(2020)}]{serrano2020political}
Serrano, J. C.~M.; et~al. 2020.
\newblock The Political Dashboard: A Tool for Online Political Transparency.
\newblock In \emph{ICWSM}.

\bibitem[{Sharma, Ferrara, and Liu(2021)}]{sharma2021characterizing}
Sharma, K.; Ferrara, E.; and Liu, Y. 2021.
\newblock Characterizing Online Engagement with Disinformation and Conspiracies
  in the 2020 US Presidential Election.
\newblock \emph{arXiv:2107.08319}.

\bibitem[{Silva et~al.(2020{\natexlab{a}})}]{silva2020facebook_ufl}
Silva, M.; et~al. 2020{\natexlab{a}}.
\newblock Facebook ad engagement in the russian active measures campaign of
  2016.
\newblock \emph{arXiv:2012.11690}.

\bibitem[{Silva et~al.(2020{\natexlab{b}})}]{silva2020facebook}
Silva, M.; et~al. 2020{\natexlab{b}}.
\newblock Facebook Ads Monitor: An Independent Auditing System for Political
  Ads on Facebook.
\newblock In \emph{WWW}.

\bibitem[{Sindhwani and Melville(2008)}]{sindhwani2008document}
Sindhwani, V.; and Melville, P. 2008.
\newblock Document-word co-regularization for semi-supervised sentiment
  analysis.
\newblock In \emph{IEEE ICDM}.

\bibitem[{Stieglitz and Dang-Xuan(2013)}]{stieglitz2013social}
Stieglitz, S.; and Dang-Xuan, L. 2013.
\newblock Social media and political communication: a social media analytics
  framework.
\newblock \emph{SNAM}.

\bibitem[{Student(1908)}]{student1908probable}
Student. 1908.
\newblock Probable error of a correlation coefficient.
\newblock \emph{Biometrika}.

\bibitem[{Subramanya and Bilmes(2008)}]{subramanya2008soft}
Subramanya, A.; and Bilmes, J. 2008.
\newblock Soft-supervised learning for text classification.
\newblock In \emph{EMNLP}.

\bibitem[{Talukdar et~al.(2008)}]{talukdar2008weakly}
Talukdar, P.; et~al. 2008.
\newblock Weakly-supervised acquisition of labeled class instances using graph
  random walks.
\newblock In \emph{EMNLP}.

\bibitem[{Tang, Qu, and Mei(2015)}]{tang2015pte}
Tang, J.; Qu, M.; and Mei, Q. 2015.
\newblock PTE: Predictive Text Embedding Through Large-scale Heterogeneous Text
  Networks.
\newblock In \emph{KDD}.

\bibitem[{Tang et~al.(2015)}]{tang2015line}
Tang, J.; et~al. 2015.
\newblock Line: Large-scale information network embedding.
\newblock In \emph{WWW}.

\bibitem[{Wattal et~al.(2010)}]{wattal2010web}
Wattal, S.; et~al. 2010.
\newblock Web 2.0 and politics: the 2008 US presidential election and an
  e-politics research agenda.
\newblock \emph{MIS quarterly}.

\bibitem[{Wu et~al.(2020)}]{wu2020clear}
Wu, Z.; et~al. 2020.
\newblock Clear: Contrastive learning for sentence representation.
\newblock \emph{arXiv:2012.15466}.

\bibitem[{Yang, Cohen, and Salakhudinov(2016)}]{yang2016revisiting}
Yang, Z.; Cohen, W.; and Salakhudinov, R. 2016.
\newblock Revisiting semi-supervised learning with graph embeddings.
\newblock In \emph{ICML}.

\bibitem[{Zhang et~al.(2020)}]{zhang2020minimally}
Zhang, Y.; et~al. 2020.
\newblock Minimally supervised categorization of text with metadata.
\newblock In \emph{ACM SIGIR}.

\bibitem[{Zhu and Ghahramani(2002)}]{Zhu02learningfrom}
Zhu, X.; and Ghahramani, Z. 2002.
\newblock Learning from Labeled and Unlabeled Data with Label Propagation.
\newblock Technical report.

\end{thebibliography}

\end{document}